\documentclass[10pt,twocolumn,letterpaper]{article}

\usepackage[pagenumbers]{cvpr}

\usepackage{graphicx}
\usepackage{amsmath}
\usepackage{amssymb}
\usepackage{booktabs}
\usepackage{comment}
\usepackage{color}
\usepackage{mathtools}
\usepackage{scalerel}
\usepackage{algorithm}
\usepackage{algpseudocode}
\usepackage{epstopdf}
\usepackage{multirow}
\usepackage{array}
\usepackage{adjustbox, colortbl}
\newcolumntype{P}[1]{>{\centering\arraybackslash}p{#1}}
\newcolumntype{M}[1]{>{\centering\arraybackslash}m{#1}}
\usepackage[page]{appendix}
\usepackage{enumitem}
\usepackage{makecell}

\usepackage{listings}

\usepackage[pagebackref,breaklinks,colorlinks]{hyperref}

\usepackage[capitalize]{cleveref}
\crefname{section}{Sec.}{Secs.}
\Crefname{section}{Section}{Sections}
\Crefname{table}{Table}{Tables}
\crefname{table}{Tab.}{Tabs.}

\def\cvprPaperID{3920} 
\def\confName{CVPR}
\def\confYear{2022}

\begin{document}

\title{\Large ML-Decoder: Scalable and Versatile Classification Head}

\author{ Tal Ridnik\thanks{Equal contribution} \hspace{0.15cm} Gilad Sharir\textsuperscript{*} \hspace{0.1cm} \\  Avi Ben-Cohen \hspace{0.1cm} Emanuel Ben-Baruch \hspace{0.1cm}  Asaf Noy
\vspace{0.5cm} \\ 
DAMO Academy, Alibaba Group\\
}

\maketitle

\begin{abstract}
    \label{abstract}
In this paper, we introduce ML-Decoder, a new attention-based classification head.  ML-Decoder predicts the existence of class labels via queries, and enables better utilization of spatial data compared to global average pooling.
By redesigning the decoder architecture, and using a novel group-decoding scheme, ML-Decoder is highly efficient, and can scale well to thousands of classes. Compared to using a larger backbone, ML-Decoder consistently provides a better speed-accuracy trade-off.
ML-Decoder is also versatile - it can be used as a drop-in replacement for various classification heads, and generalize to unseen classes when operated with word queries. Novel query augmentations further improve its generalization ability.
Using ML-Decoder, we achieve state-of-the-art results on several classification tasks:
on MS-COCO multi-label, we reach $91.4\%$ mAP; on NUS-WIDE zero-shot, we reach $31.1\%$ ZSL mAP; and on ImageNet single-label, we reach with vanilla ResNet50 backbone a new top score of $80.7\%$, without extra data or distillation. Public code is available at: \url{ https://github.com/Alibaba-MIIL/ML_Decoder}
\end{abstract}

\section{Introduction}
\label{introduction} 
\begin{figure}[t!]
\centering
\includegraphics[scale=.35]{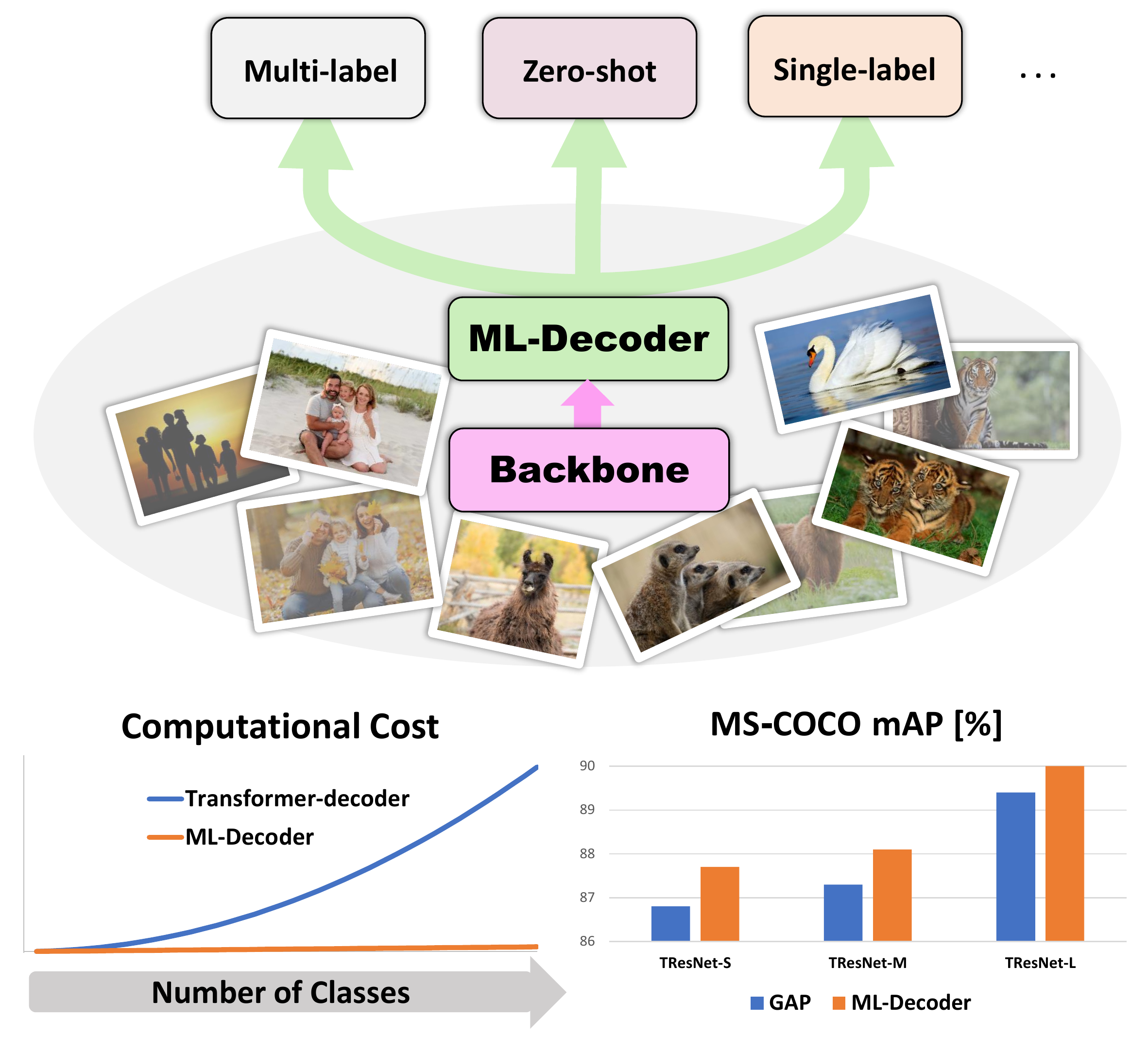}
\caption{\textbf{Our proposed classification head.} ML-Decoder is versatile, and provides a unified solution for several classification tasks, with state-of-the-art results. Unlike transformer-decoder, it is also scalable, and can handle efficiently thousands of classes.}
\label{fig:promotional_pic}
\vspace{-0.2cm}
\end{figure}

Image classification is
a vital computer-vision task, that requires assigning a label or multiple labels to an image, according to the objects present in it.
With \emph{single-label classification}~\cite{wei2014cnn, yu2013largescale}, we assume that the image contains only one object, hence we can apply a softmax operation on the output logits. However, natural images usually contain multiple objects and concepts, highlighting the importance of \emph{multi-label classification}~\cite{yang2016exploit,tsoumakas2007multi}, where we predict each class separately and independently, in a similar fashion to multi-task problems \cite{caruana1997multitask,ruder2017overview}.
Notable success in the field of multi-label classification was reported by exploiting label correlation via graph neural networks~\cite{chen2019multi_MLGCN, chen2019multi}, and improving loss functions, pretrain methods and backbones~\cite{ben2020asymmetric,ridnik2021imagenet21k,ridnik2021tresnet,benbaruch2021multilabel}.

In a regime of \emph{extreme classification}~\cite{zhang2018deep,medini2019extreme},
we need to predict the existence of a large number of classes (usually thousands or more), forcing our model and training scheme to be efficient and scalable.
Multi-label \emph{zero-shot learning} (ZSL)~\cite{xian2017zero,wang2019survey} is an extension of multi-label classification, where during inference the network tries to recognize unseen labels, i.e., labels from additional categories that were not used during training.  
This is usually done by sharing knowledge between the seen classes (that were used for training) and the unseen classes via a text model \cite{goldberg2014word2vec, pennington2014glove}.

\begin{figure*}[t!]
\centering
\includegraphics[width=0.72\paperwidth]{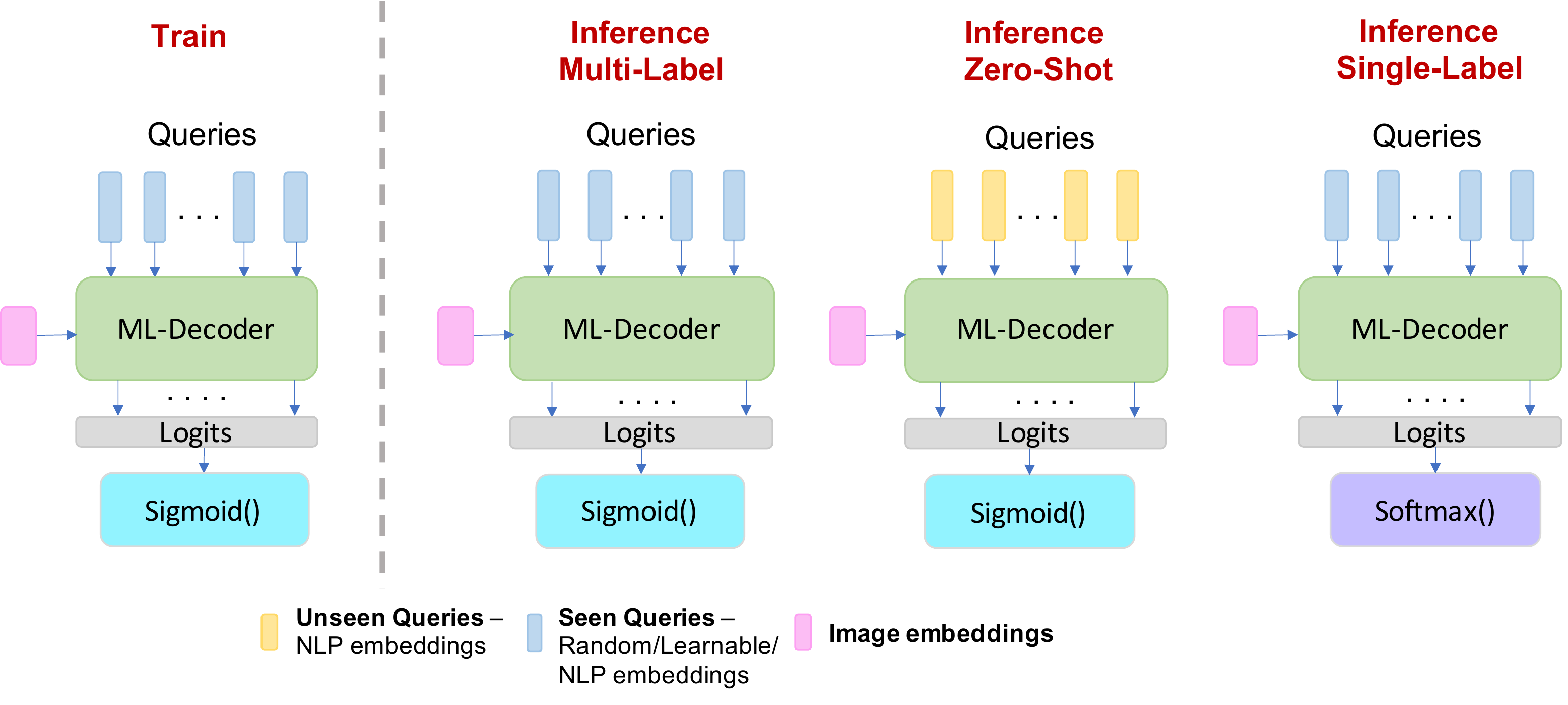}
\caption{\textbf{Versatility - } ML-Decoder module is applicable to various classification tasks, such as multi-label, zero-shot, and single-label.}
\label{fig:zsl_scheme}
\vspace{-0.2cm}
\end{figure*}

%
Classification networks usually contain a backbone, and a classification head~\cite{ridnik2021tresnet,tan2019efficientnet,he2016deep}. The backbone outputs a spatial embedding tensor, and the classification head transforms the spatial embeddings into prediction logits.
In single-label classification, this is commonly done by global-average-pooling (GAP) , followed by a fully connected layer~\cite{ghosh2018adgap}. %
GAP-based heads are also used also for multi-label classification \cite{chen2019multi_MLGCN,wu2020distribution,durand2019learning}. 
However, the need to identify several objects, with different locations and sizes, can make the usage of average pooling sub-optimal. %
Recently, several works proposed attention-based heads for multi-label classification. \cite{gao2021learning} offered a two-stream attention framework to recognize multi-category objects from
global image to local regions. \cite{zhu2021residual} suggested simple spatial attention scores, and then
combined them with class-agnostic average pooling features. \cite{liu2021query2label} presented a pooling transformer with learnable queries for multi-label classification, achieving top results.

GAP-based classification heads are simple and efficient, and scale well with the number of classes, since they have a fixed spatial pooling cost. However, they provide sub-optimal results, and are not directly applicable to ZSL.
Attention-based classification heads do improve results, but are often costly, even for datasets with a small number of classes, and practically infeasible to extreme classification scenarios. They also have no natural extension to ZSL.

In this paper, we introduce a new classification head, called \textit{ML-Decoder}, that provides a unified solution for single-label, multi-label, and zero-shot  classification, with state-of-the-art results (see Figure \ref{fig:promotional_pic}).
ML-Decoder design is based on the original transformer-decoder~\cite{vaswani2017attention}, with two major modifications, that significantly improve its scalability and efficiency. First, it reduces the quadratic dependence of the decoder in the number of input queries to a linear one, by removing the redundant self-attention block.  Second, ML-Decoder uses a novel \textit{group-decoding} scheme, where instead of assigning a query per class, it uses a fixed number of queries, that are interpolated to the final number of classes via a new architectural block called \emph{group fully-connected}. Using group-decoding, ML-Decoder also enjoys a fixed spatial pooling cost, and scales well to thousands of classes.


ML-Decoder is flexible and efficient. It can be trained equally well with learnable or fixed queries, and can use different queries during training and inference (see Figure \ref{fig:zsl_scheme}). These  key features make ML-Decoder suitable for ZSL tasks. When we assign a query per class and train ML-decoder with word queries, it generalizes well to unseen queries, and significantly improves previous state-of-the-art ZSL results. We also show that the group-decoding scheme can be extended to the ZSL scenario, and introduce novel query augmentations during training to further encourage generalization.
%



%

The paper’s contributions can be summarized as follows:
\begin{itemize}[leftmargin=0.4cm]
  \setlength{\itemsep}{3.4pt}
  \setlength{\parskip}{0.2pt}
  \setlength{\parsep}{0.2pt}
  \item We propose a new classification head called ML-Decoder, which provides a \textit{unified solution} for multi-label, zero-shot, and single-label classification, with state-of-the-art results.
  \item ML-Decoder can be used as a drop-in replacement for global average pooling. It is \textit{simple} and \textit{efficient}, and provides improved speed-accuracy trade-off compared to larger backbones, or other attention-based heads.
\item ML-Decoder novel design makes it \textit{scalable} to classification with thousands of classes.  Complementary query-augmentation technique improves its \textit{generalizability} to unseen classes as well.
\item We verify the effectiveness of  ML-Decoder
with comprehensive experiments on commonly-used classification datasets: MS-COCO, Open Images, NUS-WIDE, PASCAL-VOC, and ImageNet.

\end{itemize}


\section{Method}
In this section, we will first review the baseline classification heads. Then we will present our novel ML-Decoder, discuss its advantages, and show its applicability to several computer-vision tasks, such as multi-label, ZSL, and single-label classification.

\subsection{Baseline Classification Heads}
A typical classification network is comprised of a backbone, and a classification head.
The network's backbone outputs a spatial embedding tensor,
$E \in \mathbb{R}^{H\times W\times D}$, and the classification head transforms the spatial embeddings tensor into N logits, $\mathbf\{l_n\}_{n=1}^{N}$ , where N is the number of classes. 
There are two baseline approaches for processing the spatial embeddings: GAP-based, and attention-based.

\textbf{GAP-based}: with a GAP-based classification head, we first reduce the spatial embeddings to a one-dimensional vector via simple global averaging operation on the spatial dimensions, outputting a vector $\mathbf{z} \in \mathbb{R}^{D\times 1}$. Then, a fully connected layer transforms the embedding vector into N output logits: $\mathbf{l} = W \mathbf{z}$, where  $W \in \mathbb{R}^{N\times D}$ is a learnable linear projections matrix. GAP is commonly used for single-label classification tasks~\cite{ridnik2021tresnet, tan2019efficientnet,he2016deep}, and has some generalizations, for example~\cite{radenovic2018fine,lee2017generalizing}. GAP was also adopted as a baseline approach for mutli-label classification ~\cite{ben2020asymmetric,wang2017multi,liu2018multi}

\textbf{Attention-based}: Unlike single-label classification, in multi-label classification several objects can appear in the image, in different locations and sizes. Several works~\cite{liu2021query2label,gao2021learning,zhu2021residual} have noticed that the GAP operation, which eliminates the spatial dimension via simple averaging, can be sub-optimal for identifying multiple objects with different sizes. Instead they suggested using attention-based classification heads, which enable more elaborate usage of the spatial data, with improved results. 
%

\begin{figure}
\centering
\includegraphics[scale=.70]{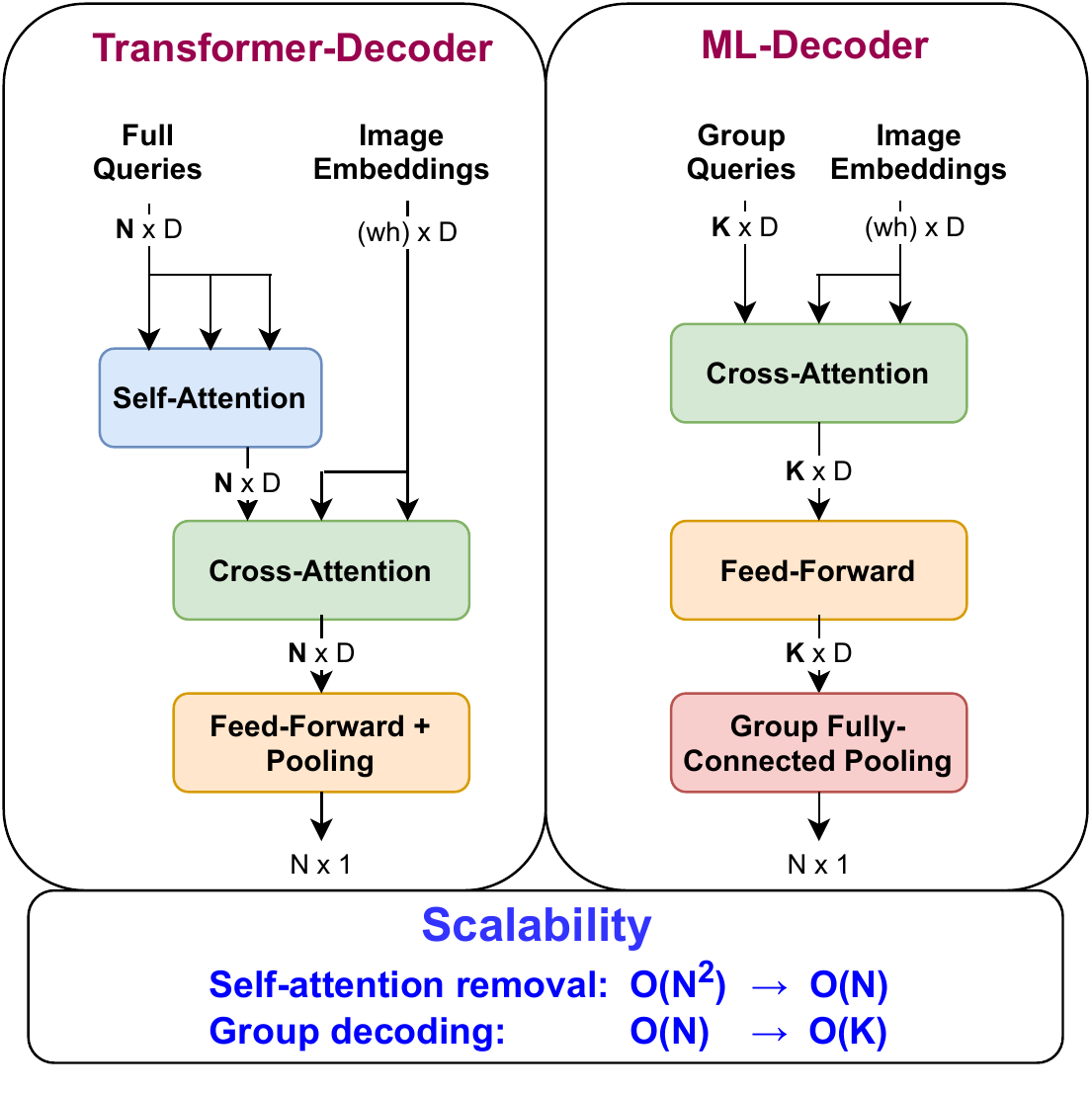}
\caption{\textbf{Scalability} - baseline transformer-decoder vs. our proposed ML-Decoder. N - number of classes, K - number of group queries, D - tokens length. Removing the redundant self-attention block relaxes the quadratic dependence in the number of queries to a linear one, while retraining the same expressivity. When using group queries with fixed number of queries $K < N$, ML-Decoder becomes fully scalable, with spatial pooling cost independent in the number of classes.}
\label{fig:intro_figure}
\vspace{-0.15cm}
\end{figure}
\subsection{Recap - Attention and Transformer-Decoder}
Among the attention-based classification heads proposed, a simple approach based on a transformer-decoder, similar to the one used by DETR for object detection~\cite{carion2020end}, has achieved top results on multi-label classification~\cite{liu2021query2label}.

A transformer-decoder unit relies on the multi-head attention module, introduced in~\cite{vaswani2017attention}. A multi-head attention has three inputs: $Q,K,V$. If we define the attention operation to be:
\begin{equation}
   \text{Attention}(Q, K, V) = \text{Softmax}(\frac{QK^T}{\sqrt{d_k}})V
\label{eq:Attention}   
\end{equation}
A multi-head module output is:
\begin{align*}
    \text{MultiHeadAttn}(Q, K, V) &= \text{Concat}(\mathrm{head_1}, ..., \mathrm{head_h})W^O\\
    \text{where}~\mathrm{head_i} &= \text{Attention}(QW^Q_i, KW^K_i, VW^V_i)
\end{align*}
$W^Q_i,W^K_i,W^V_i,W^O$ are learnable  projections matrices.

An illustration of a transformer-decoder classification head is given in Figure \ref{fig:intro_figure} (left side).
The transformer-decoder has two inputs: The spatial embedding tensor, $E$, and a set of N learnable queries, $Q$, one for each class. The transformer-decoder processes the inputs via four consecutive stages called self-attention, cross-attention, feed-forward and token-pool:
\begin{alignat}{2}
    &\text{self-attn:} &&~~Q_1 \xleftarrow{} \text{MultiHeadAttn}(Q, Q, Q) \cr
    &\text{cross-attn:}\quad &&~~Q_2 \xleftarrow{} \text{MultiHeadAttn}(Q_1, E, {E}) \cr
    &\text{feed-forward:} &&~~Q_3 \xleftarrow{} \text{FF}(Q_2)  \cr
    &\text{token-pool:} &&~~\text{Logits} \xleftarrow{} \text{Pool}(Q_3)
\label{eq:transformer-decoder}
\end{alignat}
FF is a feed-forward fully connected layer, as defined in~\cite{vaswani2017attention}.
The token pooling stage is a simple pooling on the token embeddings' dimension D, to produce $N$ output logits.

\subsection{ML-Decoder}
\subsubsection{Motivation}
On multi-label datasets with small number of classes, such as MS-COCO~\cite{lin2014microsoft} and Pascal-VOC~\cite{everingham2007pascal} ($80$ and $20$ classes respectively),
transformer-decoder classification head works well, and achieves state-of-the-art results~\cite{liu2021query2label}, with small additional computational overhead. However, it suffers from a critical drawback - the computational cost is quadratic with the number of classes. Hence, for datasets with  large number of classes, such as Open Images~\cite{kuznetsova2018open} ($9600$ classes), using transformer-decoder is practically infeasible in terms of computational cost, as we will show in Section \ref{sec:comparing_pooling_schemes}. For real-world applications, a large number of classes is imperative to provide a complete and comprehensive description of an input image. Hence, a more scalable and efficient attention-based classification head is needed. 
In addition, transformer-decoder as a classification head is suitable for multi-label classification only. A more general attention-based  head, that can also address other tasks such as single-label and ZSL, will be beneficial.

\subsubsection{ML-Decoder Design}\label{ml-decoder-design}
We will now describe our proposed classification head, ML-Decoder. Illustration of ML-Decoder flow is given in Figure \ref{fig:intro_figure} (right-side). 
Compared to transformer-decoder, ML-Decoder includes the following modifications:
\vspace{-0.2cm}
\paragraph{(1) Self-attention removal:} We start by observing that during inference, the self-attention module of transformer-decoder provides a \textit{fixed} transformation on the input queries. However, as the queries are entering the cross-attention module, they are subjected to a projection layer, before going through the attention operation (Eq.~\ref{eq:Attention}). In practice, the projection layer can transform the queries to any desired output, making the self-attention module redundant. Hence, we can remove the self-attention layer, while still maintaining the same expressivity of the classification head, and without degrading the results. We will validate this empirically in Section \ref{sec:query_typs}. By removing the self-attention, we avoid a costly module, and relax the quadratic dependence of ML-Decoder in the number of input queries to a linear one, making it more practical and efficient. 
\vspace{-0.2cm}
\paragraph{(2) Group-decoding:}
In an extreme classification scenario, even linear dependency of the classification head with the number of classes can be costly.
We want to break this coupling, and make the cross-attention module, and the feed-forward layer after it, independent of the number of classes, same as GAP operation. To this end, instead of assigning a query per class, we use as inputs a fixed number of group queries, K (see Figure \ref{fig:intro_figure}). 

After the feed-forward layer, we transform the group queries into output logits via a novel layer called \emph{group fully-connected}. This layer performs simultaneously two tasks - (1) expand each group query to $\frac{N}{K}$  outputs; (2) pool the embeddings' dimension. 
%
%
If we define the \emph{group-factor} to be $g=\frac{N}{K}$, group fully-connected generates an output logit $L_i$ with the following operation:
\begin{gather}
    L_i=(W_k\cdot Q_k)_j \nonumber \\
    \text{where:~~~}\mathrm{k}=i~~\mathbf{div}~~g,~~~\mathrm{j}=i~~\mathbf{mod}~~g 
\label{eq:group_fc}
\end{gather} 
$Q_k\in \mathbb{R}^D$ is the $k^{th}$ query, and $W_k\in \mathbb{R}^{g \times D}$ is the $k^{th}$ learnable projection matrix.
An illustration of group fully-connected layer is given in Figure \ref{fig:group_pooling}, and a pseudo-code appears in appendix \ref{appendix:group_pseudo_code}.
%
\begin{figure}[hbt!]
\centering
\includegraphics[scale=.75]{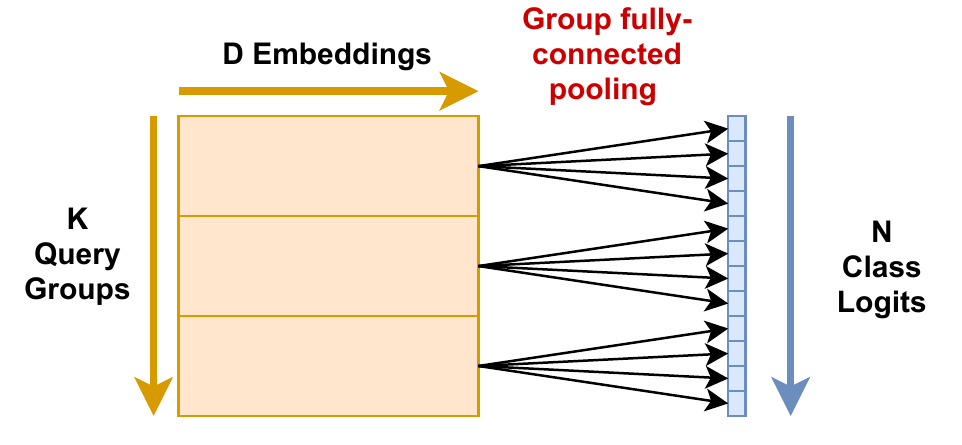}
\caption{\textbf{Scheme of a group fully-connected layer}  (with $g=4$).
}
\label{fig:group_pooling}
\vspace{-0.15cm}
\end{figure}
\\
The full flow of ML-Decoder with group-decoding is depicted in Eq. \ref{eq:ML-decoder}, where $G_q$ are the input group queries:
\begin{alignat}{2}
    &\text{cross-attn:}\quad &&~~G_{q_1} ~~~~\xleftarrow{} \text{MultiHeadAttn}(G_q, E, {E}) \cr
    &\text{feed-forward:} &&~~G_{q_2} ~~~~\xleftarrow{} \text{FF}(G_{q_1}) \cr
    &\text{group FC:} &&~~\text{Logits} \xleftarrow{} \text{Group-FC}(G_{q_2})
\label{eq:ML-decoder}
\end{alignat}
\\
Some additional observations and insights into the group-decoding scheme:
\begin{itemize}[leftmargin=0.3cm]
  \setlength{\itemsep}{1.0pt}
  \setlength{\parskip}{0.2pt}
  \setlength{\parsep}{0.2pt}
\item With full-decoding ($g=1$), each query checks the existence of a single class. With group-decoding, each query checks the existence of several classes.
We chose to divide the classes into groups in a random manner. Clustering the classes via semantic proximity is an alternative, but will require a cumbersome clustering process, with extra hyper-parameters that might need tuning per dataset. In Section \ref{sec:comparing_pooling_schemes} we will show that random group clustering is enough to provide results comparable to a full-decoding scheme.
\item In terms of flops, the group fully-connected layer is equivalent to a fully-connected layer in a GAP-based head ($N \times D$ multiplications). Both are linearly dependent in the number classes, but in practice they have a small computational overhead, even for thousands of classes. 
\\In terms of memory consumption, performing the two tasks of group fully-connected together, in a single operation, is more efficient than doing them consecutively, since there is no need to store large intermediate maps.
\item The only component in ML-Decoder that depends on the input image size is the cross-attention module. We can think of the cross-attention layer as doing \emph{spatial pooling}, similar to GAP. With group-decoding, ML-Decoder has fixed spatial pooling cost, independent of N.
\end{itemize}
\vspace{-0.2cm}
\paragraph{(3) Non-learnable queries:}
\cite{liu2021query2label} argued that transformer-decoder for multi-label classification achieves top results only with learnable queries. However, we observe that the queries are always fed into a multi-head attention layer, that applies a learnable projection on them (Eq.~\ref{eq:Attention}). Hence, setting the queries weights as learnable is redundant -
a learnable projection can transform any fixed-value query to any value obtained by a learnable query. 
We will validate this empirically in Section~\ref{sec:query_typs}, showing that the same accuracies are obtained when training ML-Decoder with learnable or fixed queries. In addition to simplifying the training process, using fixed queries will enable us to do ZSL.

\subsubsection{ML-Decoder for ZSL}\label{scheme_zsl}
Next, we will present the adaptations needed in order to use ML-Decoder in a multi-label ZSL scenario, and discuss key features of ML-Decoder that make it suitable for the task. We will also show that the group-decoding scheme can be extended to ZSL, and present novel query augmentations that further improve ML-Decoder generalizability.
\vspace{-0.1cm}
\paragraph{NLP-based queries:} 
We begin by presenting a version of ML-Decoder for ZSL with a full-decoding scheme (each label has a corresponding query).
As discussed in the previous section, the input queries can be either learnable or fixed. For ZSL, we use fixed NLP-based queries - for each label, a word embedding vector is extracted using a language model, and set as the input query. 
We also use shared projection matrix in the group fully-connected layer (setting $W_k=W$ in Eq.~\ref{eq:group_fc}).
With NLP-based queries and a shared projection matrix,
semantic information can propagate from the \emph{seen} (training) classes to the \emph{unseen} (test) classes during inference, enabling generalization. 
%
\paragraph{ML-Decoder features:} ML-Decoder contains several favorable features which make it well suited for ZSL. Firstly, its attention mechanism is based on dot-product similarity between vectors (Eq.~\ref{eq:Attention}). Since NLP word embeddings preserve this dot-product semantic similarity~\cite{goldberg2014word2vec}, the unseen labels are more likely to be matched with the most similar keys and values within the decoder. 
In addition, ML-Decoder with a shared projection matrix allows a variable number of input queries, and is not sensitive to the order of queries. This is beneficial since in ZSL we perform training and testing on different sets of classes, and therefore different sets of queries. 
For ZSL we train exclusively on the seen labels, and perform inference on the unseen classes, while for Generalized ZSL (GZSL), we perform inference on the union of the unseen and seen sets of labels. 
\paragraph{Group-decoding:} Group-decoding (with $K<N$) requires modifications in order to work in a ZSL setting. In appendix~\ref{zsl_scalability} we thoroughly detail our variant of group-decoding for ZSL. 
%
%
\vspace{-0.2cm}
\paragraph{Query augmentations} 
With NLP-based queries, ML-Decoder naturally extends to the task of ZSL. Yet, we want to apply dedicated training tricks that further improve its generalizability.
It is a common practice in computer-vision to apply augmentations on the input images to prevent overfitting, and improve generalizability to new unseen images. Similarly, we introduce \emph{query augmentations} to encourage generalization to unseen class queries.
The first augmentation is \emph{random-query}, which adds additional random-valued queries to the set of input queries, and assigns a positive ground truth label signifying “random” for these added queries. The second augmentation is \emph{query-noise}, where we add each batch a small random noise to the input queries. See Figure~\ref{fig:query_aug} in the appendix for illustration of the augmentations. In  Section~\ref{sec:zsl_results} we will show that query augmentations encourage the model to recognize novel query vectors which it hasn’t encountered before, and improve ZSL scores. We also tried query-cutout augmentation, where random parts of the queries are deleted each batch. However, this technique was not beneficial in our experiments. 

\subsubsection{ML-Decoder for Single-label Classification}
Most attention-based classification heads previously proposed were purposed for multi-label classification (\cite{zhu2021residual,liu2021query2label,gao2021learning}, for example), and this was also the primary task we focused on in our work.
However, our design of ML-Decoder enables it to be used as a drop-in replacement for GAP-based heads on other computer-vision tasks, such as single-label classification, as shown in Figure \ref{fig:zsl_scheme}.

The main motivation for attention-based heads in multi-label classification was the need to identify several objects, with different locations and sizes~\cite{liu2021query2label}. We will show in Section \ref{sec:single_label_results} that the benefit from ML-Decoder is more general, and fully applies also to single-label problems, where the image usually contains a single object.


\section{Experimental Study}
\label{experiments}
In this section, we will bring ablation tests and inner comparisons for our proposed ML-Decoder classification head.
First, we will test ML-Decoder with different types of input queries. Then we will compare ML-Decoder to other classification heads, such as transformer-decoder and GAP-based. Finally, we will provide an ablation study for ZSL with augmentation queries and group-decoding.

\subsection{Comparing Query Types}
\label{sec:query_typs}
As discussed in Section \ref{ml-decoder-design}, due to the linear projection in the attention module, ML-Decoder retains the same expressivity when it uses learnable or fixed queries.
In Table \ref{Table:coco_query_types} in the appendix we compare results for ML-Decoder with different types of queries, on MS-COCO multi-label dataset (see appendix \ref{appendix:training_details_coco} for full  training details on MS-COCO).

Indeed we see that learnable, fixed random, and fixed NLP-based word queries all lead to the same accuracy, $88.1\%$ mAP, as expected. For reducing the number of learned parameters, we shall use fixed queries.

\subsection{Comparing Different Classification Heads}
\label{sec:comparing_pooling_schemes}
In Table \ref{Table:coco_decoder_schems}  we compare MS-COCO  results for training with different classification heads.
\begin{table}[hbt!]
\centering
\begin{tabular}{c|c|c|c|c} 
\Xhline{3\arrayrulewidth}
\begin{tabular}[c]{@{}c@{}}Classification \\Head\end{tabular}   & \begin{tabular}[c]{@{}c@{}}Num of\\Classes\end{tabular} & \begin{tabular}[c]{@{}c@{}}Num of \\Queries\end{tabular} & \begin{tabular}[c]{@{}c@{}}Flops \\{[}G]\end{tabular} & \begin{tabular}[c]{@{}c@{}}mAP \\{[}\%]\end{tabular}  \\ 
\Xhline{3\arrayrulewidth}
GAP                                                        & 80                                                         & —                                                           & 23.0                                                  & 87.0                                                  \\ 
\hline
\begin{tabular}[c]{@{}c@{}}Transformer- \\Decoder\end{tabular} & 80                                                         & 80                                                          & 24.1                                                   & \textbf{88.1}                                                  \\ 
\hline
ML-Decoder                                                 & 80                                                         & 20                                                          & 23.6                                                   & 88.0                                                  \\
ML-Decoder                                                 & 80                                                         & 80                                                          & 23.9                                                & \textbf{88.1}                                                  \\
\hline
\end{tabular}
\caption{\textbf{Comparison of multi-label MS-COCO mAP score for different classification heads}. Architecture - TResNet-M~\cite{ridnik2021tresnet}. 
}
\label{Table:coco_decoder_schems}
\vspace{-0.15cm}
\end{table}
\\
From Table \ref{Table:coco_decoder_schems} we learn the following observations:
\begin{itemize}[leftmargin=0.4cm]
  \setlength{\itemsep}{2.0pt}
  \setlength{\parskip}{0.2pt}
  \setlength{\parsep}{0.2pt}
  \item ML-Decoder (and transformer-decoder) provide a significant improvement (more than $1\%$ mAP), compared to GAP-based classification head.
  \item When using the same number of input queries ($80$), transformer-decoder and ML-Decoder reach the same accuracy, demonstrating that indeed the self-attention module provides a redundant transformation (see Section \ref{ml-decoder-design}), and removing it in ML-Decoder reduces the computational cost without impacting the results.
  \item Using group decoding with a ratio of $\frac{N}{K}=4$ has a minimal impact on the results - a reduction of only $0.1\%$ mAP. However, since MS-COCO dataset has a small number of classes, the additional flops from an attention-based classification head are minimal, so reducing the number of queries is not essential in this case.
\end{itemize}

In Table \ref{Table:open_images_decoder_schems} we repeat the same comparison on Open Images multi-label dataset, which has significantly more classes - $9600$ instead of $80$. Full training details on Open Images are given in appendix \ref{appendix:training_details_open_images}.

\begin{table}[hbt!]
\centering
\begin{tabular}{c|c|c|c|c} 
\Xhline{3\arrayrulewidth}
\begin{tabular}[c]{@{}c@{}}Classification\\Head\end{tabular}    & \begin{tabular}[c]{@{}c@{}}Num of\\Classes\end{tabular} & \begin{tabular}[c]{@{}c@{}}Num of \\Queries\end{tabular} & \begin{tabular}[c]{@{}c@{}}Flops \\{[}G]\end{tabular} & \begin{tabular}[c]{@{}c@{}}mAP \\{[}\%]\end{tabular}  \\ 
\Xhline{3\arrayrulewidth}
GAP                                                        & 9600                                                       & —                                                           & 5.8                                                   & 86.0                                                  \\ 
\hline
\begin{tabular}[c]{@{}c@{}}Transformer \\Decoder\end{tabular} & 9600                                                       & 9600                                                        & 178.6                                                 & NA                                                    \\ 
\hline
ML-Decoder                                                 & 9600                                                       & 100                                                         & 6.3                                                   & 86.7                                                  \\
ML-Decoder                                                 & 9600                                                       & 200                                                         & 6.7                                                   &   \textbf{86.8}                                                   \\

ML-Decoder                                                 & 9600                                                       & 400                                                        & 7.6                                                  &    \textbf{86.8}                                                   \\
\hline
\end{tabular}
\caption{\textbf{Comparison of Open Images mAP score for different classification heads.} Architecture - TResNet-M.  
}
\label{Table:open_images_decoder_schems}
\vspace{-0.15cm}
\end{table}

On Open Images, using transformer-decoder classification head is not feasible - due to the large number of classes, the additional computational cost is very high, and even with a batch size of $1$ and input resolution of $224$, our training is out-of-memory (see Table \ref{Table:througput_measurements} for full specifications). In contrast, ML-Decoder with group-decoding increases the flops count by only $10\%-20\%$, while significantly improving the mAP score on this challenging extreme classification dataset, compared to GAP. We also see from Table \ref{Table:open_images_decoder_schems} that group decoding with a ratio of $\frac{N}{K}=48$ already gives the full score benefit, and further increasing the ratio to  $\frac{N}{K}=96$ reduces the score by only $0.1\%$.

\begin{figure}[hbt!]
\centering
\includegraphics[scale=.75]{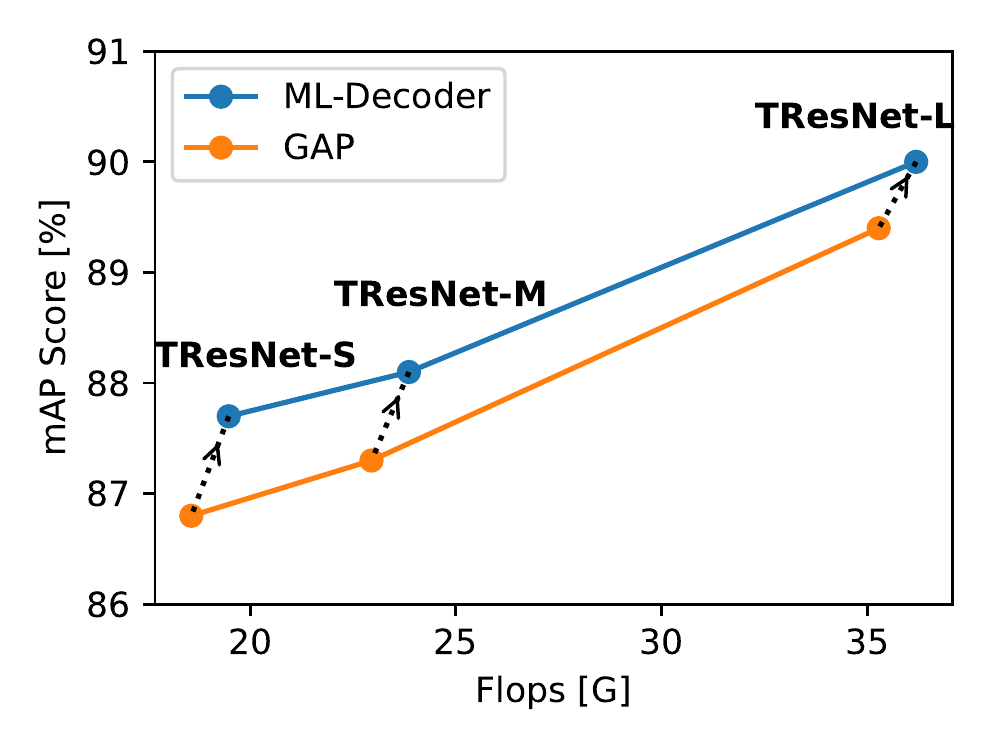}
\vspace{-0.15cm}
\caption{\textbf{mAP score vs. Flops comparison, on MS-COCO datase, for different classification heads.}
For ML-Decoder, we used $K=N=80$.}
\label{fig:flops_vs_map_coco}
\vspace{-0.15cm}
\end{figure}

In Figure \ref{fig:flops_vs_map_coco} we compare on MS-COCO the mAP score vs. flops for GAP-based  and ML-Decoder classification head, with three different architectures - TResNet-S, TResNet-M, TResNet-L (equivalent in runtime to ResNet34, ResNet50 and ResNet101~\cite{ridnik2021tresnet}).
\\
We see from Figure \ref{fig:flops_vs_map_coco} that using ML-decoder provides a better flops-accuracy trade-off compared to using GAP-based classification head with a larger backbone.

In addition to the flops-accuracy measurements, in Table \ref{Table:througput_measurements} in the appendix we provide full speed-accuracy comparisons of different classification heads, measuring inference speed, training speed, maximal batch size and flops. This table can aid in future comparisons of our work for different speed-accuracy metrics.

\subsection{Zero-shot Learning}
\label{sec:zsl_results}
This section presents an ablation study of ML-Decoder for ZSL, on NUS-WIDE dataset~\cite{chua2009nus}. NUS-WIDE is the most widely used benchmark for the multi-label ZSL task, and therefore we focus on this dataset. 
It is comprised of $925$ seen labels, and $81$ unseen labels.
%
Full training and dataset details are given in appendix \ref{appendix:training_details_nus_wide}.

In Table~\ref{Table:query_augmentation} we compare the different types of query augmentations presented in Section~\ref{scheme_zsl}.
\begin{table}[hbt!]
\centering
\begin{tabular}{c|c} 
\Xhline{3\arrayrulewidth}
\begin{tabular}[c]{@{}c@{}}Augmentation \\Type\end{tabular} & \begin{tabular}[c]{@{}c@{}}mAP {[}\%] \\(ZSL)\end{tabular}   \\ 
\Xhline{3\arrayrulewidth}
None                           & 29.9             \\ 
Additive noise                          & 30.6          \\
Random-query                            & 30.7         \\
Both                                    & \textbf{31.1}         \\
\hline
\end{tabular}
\caption{\textbf{Comparison of NUS-WIDE ZSL mAP scores for ML-Decoder with different query augmentations}. 
\vspace{-0.15cm}
}
\label{Table:query_augmentation}
\end{table}
\\
It is evident from the table that both random-query and additive noise contribute to the model's ability to generalize to unseen classes. When applying both of them, we see an increase of $1.2\%$ in the mAP score of the unseen classes. 
%
We also tested the impact of query augmentations on the seen classes. Both on MS-COCO and on NUS-WIDE, adding query augmentations had no impact on the seen classes' mAP score ($88.1\%$  on MS-COCO,  $22.7\%$ on NUS-WIDE). This is in agreement with our results in Section \ref{sec:query_typs}, where learnable and fixed queries provided the same results on the seen classes. The benefit from using query augmentation is by better generalization to the unseen classes.

In Table~\ref{Table:zsl_group_scheme} we compare group-decoder to full-decoder for the ZSL task. The modifications to the group-decoding scheme for ZSL are described in appendix~\ref{zsl_scalability}.
\begin{table}[hbt!]
\centering
\begin{tabular}{c|c|c|c} 
\Xhline{3\arrayrulewidth}
\begin{tabular}[c]{@{}c@{}}Classification \\Head\end{tabular} & \begin{tabular}[c]{@{}c@{}}Num of \\Classes\end{tabular} & \begin{tabular}[c]{@{}c@{}}Num of \\Queries\end{tabular} & \begin{tabular}[c]{@{}c@{}}mAP {[}\%] \\(ZSL)\end{tabular}  \\ 
\Xhline{3\arrayrulewidth}
ML-Decoder                                                    & 1006                                                     & 1006                                                     & 31.1                                               \\
ML-Decoder                                                    & 1006                                                     & 100                                                      & 28.7                                               \\
\hline
\end{tabular}
\caption{\textbf{Comparison of NUS-WIDE ZSL mAP scores for ML-Decoder with different number of input queries}. 
\vspace{-0.15cm}
}
\label{Table:zsl_group_scheme}
\end{table}

We see from the Table~\ref{Table:zsl_group_scheme} that the group-decoding scheme works well also for the ZSL scenario, with a small decrease in the mAP scores compared to full-decoding.



\section{Results}
\label{benchmark results}
In this section, we will evaluate our ML-Decoder-based solution on popular multi-label, ZSL, and single-label classification datasets, and compare results to known state-of-the-art techniques.

\subsection{Multi-label Classification}
\subsubsection{MS-COCO}
MS-COCO \cite{lin2014microsoft} is a commonly-used dataset to
evaluate multi-label image classification. It contains $122,218$ images from $80$ different categories,  divided to a training set of $82,081$ images and a validation set of $40,137$ images.

In Table \ref{Table:coco_full_comparison} we compare ML-Decoder results to top known solutions from the literature. Full training details appear in appendix \ref{appendix:training_details_coco}.
\begin{table}[hbt!]
\centering
\begin{tabular}{c|c|c|c} 
\Xhline{3\arrayrulewidth}
Method     & Backbone  & \begin{tabular}[c]{@{}c@{}}Input \\Resolution\end{tabular} & \begin{tabular}[c]{@{}c@{}}mAP \\{[}\%]\end{tabular}  \\ 
\Xhline{3\arrayrulewidth}
ML-GCN~\cite{chen2019multi_MLGCN}     & ResNet101 & 448x448                                                    & 83.0                                                  \\ 
KSSNET~\cite{liu2018multi}     & ResNet101 & 448x448                                                    & 83.7                                                  \\ 
SSGRL~\cite{chen2019learning}     & ResNet101 & 576x576                                                    & 83.8                    \\

MS-CMA~\cite{you2020cross}     & ResNet101 & 448x448                                                    & 83.8                                                  \\ 
ASL~\cite{ben2020asymmetric}        & ResNet101 & 448x448                                                    & 85.0                                                  \\
ASL~\cite{ben2020asymmetric}        & TResNet-L & 448x448                                                    & 88.4                                                  \\ 
Q2L~\cite{liu2021query2label}        & TResNet-L & 448x448                                                    & 89.2                                                  \\ 
\hline
ML-Decoder & ResNet101 & 448x448                                                    & 87.1                                                   \\
ML-Decoder & TResNet-L & 448x448                                                    & \textbf{90.0}                                         \\

\hline
\end{tabular}
\caption{\textbf{State-of-the-art comparison on MS-COCO dataset.} TResNet-L has equivalent runtime to ResNet101, with design tricks that lead to improved results \cite{ridnik2021tresnet}.}
\label{Table:coco_full_comparison}
\end{table}

Since previous works did not always report their computational cost, and released a reproducible code, we cannot provide a full speed-accuracy comparison to them. Still, we see that with ML-Decoder we improve results on MS-COCO dataset, with minimal additional computational cost compared to plain GAP-based solution (see also Table \ref{Table:coco_decoder_schems}). We hope that future works will use our results as a baseline for a more complete comparison, including computational cost and accuracy. For that matter, in Table \ref{Table:coco_resolution_comparison} in the appendix we report our results and flops count for different input resolutions. Note that with input resolution of $640$, we reach on MS-COCO with TResNet-L and TResNet-XL new state-of-the-art results of $91.1\%$ and $91.4\%$, respectively.

\subsubsection{Additional Multi-label Datasets}
\textbf{Pascal-VOC:} In Table \ref{Table:pascal_voc} in the appendix we present results on another popular multi-label dataset - Pascal-VOC~\cite{everingham2007pascal}. With ML-Decoder we reach new state-of-the-art result on Pascal-VOC - $96.6\%$ mAP. \\
\textbf{Open Images:}
In Table \ref{Table:open_images} in the appendix we present results on an extreme classification multi-label dataset - Open Images~\cite{kuznetsova2018open}, which contains $9600$ classes. Also on this dataset ML-Decoder outperforms previous methods, achieving $86.8\%$ mAP. Notice that due the large number of classes, some attention-based methods are not feasible for this dataset. Hence, no results for them are available.

\subsection{Zero-Shot Learning}

In Table~\ref{tab:sota_nuswide} we present a SotA comparison on NUS-WIDE multi-label zero-shot dataset~\cite{chua2009nus}. Similar to previous works, we use F1 score at top-K predictions and mAP as evaluation metrics. mAP is measured both on ZSL (unseen classes only) and GZSL (seen+unseen classes). Full training details appear in appendix \ref{appendix:training_details_nus_wide}.
\begin{table}[hbt!]
\centering

\adjustbox{width=\linewidth}{
\begin{tabular}{ccccc} 
\toprule[0.15em]
\multirow{1}{*}{\textbf{ Method}} & \multirow{1}{*}{\textbf{Task }} & \multirow{1}{*}{\textbf{mAP }} & \multicolumn{1}{c}{\textbf{F1 (K = 3) }} & \multicolumn{1}{c}{\textbf{F1 (K = 5) }}   \\
\toprule[0.15em]
\multirow{2}{*}{CONSE~\cite{norouzi2013zero}} & ZSL & 9.4 & 21.6 & 20.2   \\
 & GZSL & 2.1 &  7.0 & 8.1  \\ 
\cmidrule(lr){2-5}
\multirow{2}{*}{Fast0Tag~\cite{zhang2016fast}} & ZSL & 15.1 & 27.8 & 26.4  \\
 & GZSL & 3.7 & 11.5 & 13.5  \\ 
\cmidrule(lr){2-5}
\multirow{2}{*}{Attention per Label~\cite{kim2018bilinear}} & ZSL  & 10.4 & 25.8 & 23.6 \\
 & GZSL & 3.7 &  10.9 & 13.2   \\ 
\cmidrule(lr){2-5}
\multirow{2}{*}{Attention per Cluster~\cite{huynh2020shared}} & ZSL & 12.9 & 24.6 &  22.9  \\
 & GZSL  & 2.6 &  6.4 & 7.7  \\ 
\cmidrule(lr){2-5}
\multirow{2}{*}{LESA~\cite{huynh2020shared}} & ZSL & 19.4 & 31.6 & 28.7  \\
 & GZSL & 5.6 & 14.4 & 16.8   \\ 
\cmidrule(lr){2-5}

\multirow{2}{*}{BiAM~\cite{narayan2021discriminative}} & ZSL  & {26.3} & 33.1 & 30.7     \\
 & GZSL & {9.3} &  {16.1}   & {19.0}     \\
 \cmidrule(lr){2-5}
\multirow{2}{*}{SDL~\cite{ben2021semantic}} & ZSL  & 25.9 & 30.5 & 27.8     \\
 & GZSL & 12.1 &  18.5   & 21.0     \\
 \cmidrule(lr){2-5}
 \multirow{2}{*}{\textbf{ML-Decoder}} & ZSL  & \textbf{31.1} & \textbf{34.1} & \textbf{30.8}     \\
 & GZSL & \textbf{19.9} &  \textbf{23.3}   & \textbf{26.1}     \\

\bottomrule[0.1em]
\end{tabular}
}
\vspace{-0.15cm}
\caption{\textbf{State-of-the-art comparison for multi-label  ZSL and GZSL tasks on NUS-WIDE dataset}. 
}
\label{tab:sota_nuswide}
\end{table}

\\
We see from Table~\ref{tab:sota_nuswide} that our approach significantly outperforms the previous top solution by $4.8\%$ mAP (ZSL), setting a new SotA in this task.

Note that previous methods were mostly aimed at optimizing the task of zero-shot, at the expense of the seen classes (ZSL vs. GZSL trade-off). SDL~\cite{ben2021semantic}, for example, proposed to use several principal embedding vectors, and trained them using a tailored loss function for ZSL.
In contrast to previous methods, ML-Decoder offers a simple unified solution for plain and zero-shot classification, and achieves top results for both ZSL and GZSL. 
ML-Decoder sets a new SotA score also on the GZSL task, outperforming SDL with a significant improvement (from $12.1\%$ to $19.9\%$). 
This clearly demonstrates that ML-Decoder generalizes well to unseen classes, while maintaining high accuracy on the seen classes.

\subsection{Single-label Classification}
\label{sec:single_label_results}
\begin{figure}[hbt!]
\centering
\includegraphics[scale=.7]{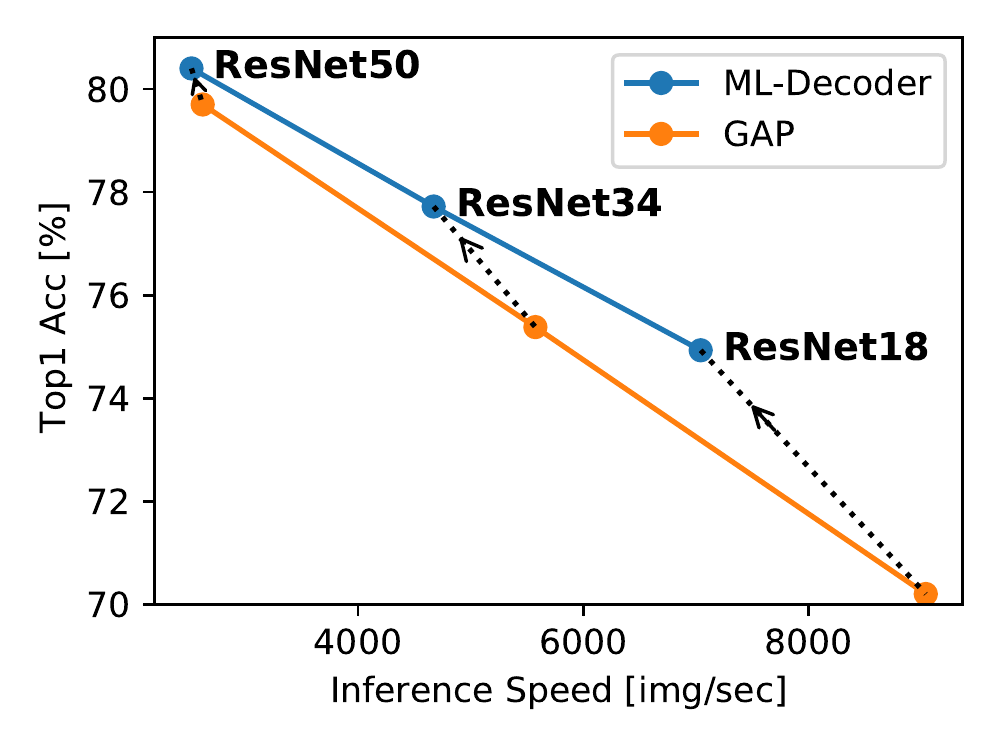}
\caption{\textbf{Speed-accuracy comparison for different classification heads on ImageNet dataset}.
For ML-Decoder, we used group-decoding (100 groups). 
Training configuration - A2~\cite{wightman2021resnet}.
}
\label{fig:resnet_inferece_score}
\vspace{-0.15cm}
\end{figure}

To test ML-Decoder effectiveness for single-label classification, we used ImageNet dataset~\cite{deng2009imagenet}, with the high-quality training code suggested in \cite{wightman2021resnet} (A2 configuration). Comparison of various ResNet architectures, with different classification heads, appears in Figure \ref{fig:resnet_inferece_score}.
As can be seen, when replacing the baseline GAP-based head with ML-Decoder, we significantly improve the models' accuracy. ML-Decoder also provides a better speed-accuracy trade-off compared to using GAP with a larger backbone.

Notice that we used ML-Decoder without introducing any change or tuning any hyper-parameter in the training configuration. The fact that ML-Decoder can serve as a drop-in replacement for GAP-based classification head, in a highly-optimized single-label training setting (baseline ResNet50 achieves $79.7\%$), and still provides an additional boost, demonstrate its effectiveness and versatility. 
Also, note that following \cite{wightman2021resnet}, our training configuration for ImageNet is using a multi-label loss (sigmoid with BCE loss instead of softmax and CE loss). That's the default mode we present in Figure \ref{fig:zsl_scheme}. In Table \ref{Table:single_label_softmax_results} in the appendix we validate that also with softmax, ML-Decoder provides a benefit compared to plain GAP.

When increasing the number of training epochs to $600$ (A1 configuration in~\cite{wightman2021resnet}), with ML-Decoder and vanilla ResNet50 backbone we reach $80.7\%$ accuracy. To the best of our knowledge, this is the top results so far achieved with ResNet50 (without extra data or distillation).

\subsubsection{Additional Results}
In appendix \ref{appendix:additional_results} we bring additional results for single-label classification, showing that with ML-Decoder we achieve top results on two other prominent single-label datasets - CIFAR-100\cite{cifar} and Stanford-Cars\cite{stanford_cars}.

\section{Conclusions and Future Work}
\label{conclusion}
In this paper, we introduced ML-Decoder, a new attention-based classification head.
By removing the redundant self-attention layer and using a novel group-decoding scheme, ML-Decoder scales well to thousands of classes, and provides a better speed-accuracy trade-off than using a larger backbone.
ML-Decoder can work equally well with fixed or random queries, and use different queries during training and inference.
With word-based queries and novel query augmentations, ML-Decoder also generalizes well to unseen classes.
Extensive  experimental  analysis  shows that ML-Decoder outperforms GAP-based heads on several classification tasks, such as multi-label, single-label and zero-shot, and achieves new state-of-the-art results.

Our future work will focus on extending the usage of ML-Decoder to other computer-vision tasks that include classification, such as object detection and video recognition. We will also explore using the group-decoding scheme more generally, for processing any spatial embedding tensor, and will try to apply it to other fields and tasks, such as segmentation, pose estimation and NLP-related problems. 


{\small
\bibliographystyle{ieee_fullname.bst}
\bibliography{egbib.bib}
}

\clearpage
\appendix
\begin{appendices}

\lstset{language=Python}
\lstset{frame=lines}
\lstset{label={lst:code_direct}}
\lstset{basicstyle=\footnotesize,columns=fullflexible}

\definecolor{codegreen}{rgb}{0,0.6,0}
\definecolor{codegray}{rgb}{0.5,0.5,0.5}
\definecolor{codepurple}{rgb}{0.58,0,0.82}
\definecolor{backcolour}{rgb}{0.95,0.95,0.92}
\lstdefinestyle{mystyle}{
    commentstyle=\color{codegreen},
    keywordstyle=\color{magenta},
    numberstyle=\tiny\color{codegray},
    stringstyle=\color{codepurple},
    basicstyle=\ttfamily\footnotesize,
    breakatwhitespace=false,         
    breaklines=false,                 
    captionpos=b,                    
    keepspaces=true,                 
    numbers=left,                    
    numbersep=5pt,                  
    showspaces=false,                
    showstringspaces=false,
    showtabs=false,                  
    tabsize=2
}
\lstset{style=mystyle}

\section{Query Types Comparison}
\begin{table}[hbt!]
\centering
\begin{tabular}{c|c} 
\Xhline{3\arrayrulewidth}
Type of Queries & mAP [\%]  \\ 
\Xhline{3\arrayrulewidth}
Learnable       & 88.1      \\ 
Fixed NLP-based             & 88.1      \\ 
Fixed Random             & 88.1      \\ 
\hline

\end{tabular}
\caption{\textbf{Comparison of MS-COCO mAP scores for ML-Decoder with different query types}. We used $K=N=80$.
}
\label{Table:coco_query_types}
\end{table}

\section{ZSL Group Decoder}
\label{zsl_scalability}
\begin{figure}[hbt!]
\centering
\includegraphics[width=0.4\textwidth]{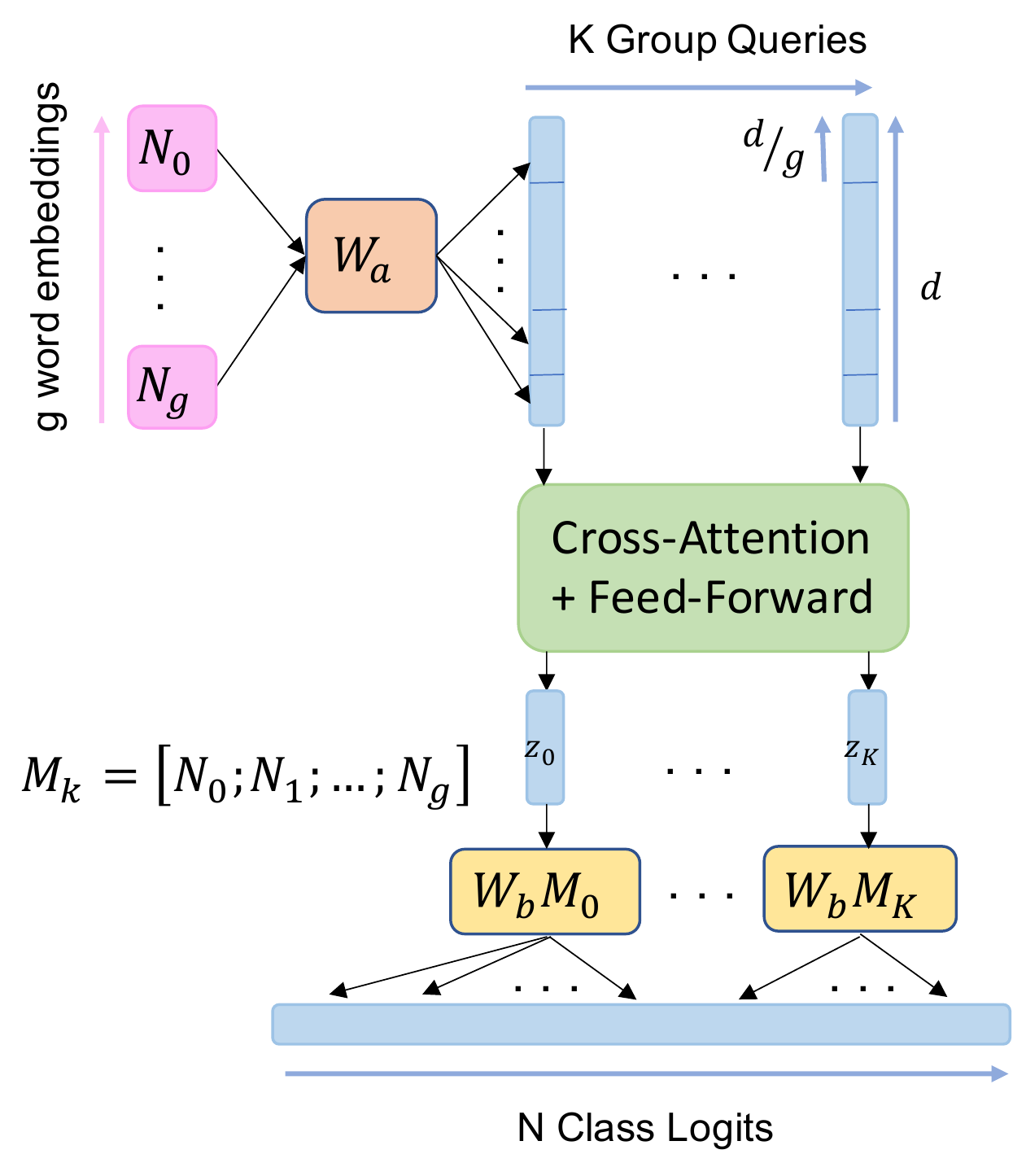}
\caption{\textbf{Group decoding scheme for ZSL.} }
\label{fig:zsl_group}
\end{figure}

We will now describe how to incorporate group-decoding in the ZSL setting in order to improve the ML-Decoder's scalability with respect to the number of classes.
Group-decoding cannot be trivially applied for ZSL, since in group-decoding each query is associated with a group of labels, while in the ZSL setting we assume that each query is associated with a specific word embedding. To alleviate this issue, we propose to construct each query by concatenating linear projections of all the word embeddings assigned to its group (See Figure~\ref{fig:zsl_group}). Formally, the $k^{th}$ group query is given by
\begin{equation}
q_k = concat(\{W_a\cdot N_i\}_{i\in \mathcal{G}_j})
\end{equation}
where $\mathcal{G}_k$ is the set of labels assigned to the $k^{th}$ group, $\{N_i\}$ is the set of word embeddings ($N_i \in \mathbb{R}^{d_w}$), $W_a\in \mathbb{R}^{\frac{d}{g}\times d_w}$ is the parameter matrix, and $g=\frac{N}{K}$.

In addition, as can be seen in Table~\ref{Table:zsl_group_ablation}, the group fully-connected head, even with shared weights, does not generalize well to unseen classes in a group-decoding setting. Therefore, we implemented a different pooling strategy for ZSL group-decoding:
%
we decompose the group fully-connected parameter matrix ($W_k$ from Eq.~\ref{eq:group_fc}) into two components: $\{N_i\}$ - the set of word embeddings which is label-specific, and $W_b\in \mathbb{R}^{d\times d_w}$ - a learned parameter matrix which is shared for all labels. 
Formally, the parameter matrix $W_k$ of the $k^{th}$ group-query (Eq.~\ref{eq:group_fc}) is constructed by the following: 
\begin{equation}
    W_k = W_b \cdot M_k 
\label{eq:parameter_w_k_zsl}
\end{equation}
where $M_k\in \mathbb{R}^{d_w \times g}$ is constructed by stacking the word embedding vectors $\{N_i\}_{i\in \mathcal{G}_k}$ of group $k$. 
Inserting Eq.~\ref{eq:parameter_w_k_zsl} into Eq.~\ref{eq:group_fc},  we get the output logits for the ZSL group-decoder.
(see Fig.~\ref{fig:zsl_group}, and pseudo-code in appendix~\ref{appendix:zsl_group_code}). Table~\ref{Table:zsl_group_ablation} shows ablation experiments for the different modifications. 
\begin{table}[hbt!]
\centering
\begin{tabular}{c|c|c} 
\Xhline{3\arrayrulewidth}
\begin{tabular}[c]{@{}c@{}}Query \\Type\end{tabular} & \begin{tabular}[c]{@{}c@{}}Head \\Type\end{tabular} & \begin{tabular}[c]{@{}c@{}}mAP {[}\%] \\(ZSL)\end{tabular}  \\ 
\Xhline{3\arrayrulewidth}
NLP               & Learnable         & 2.4          \\ 
NLP               & Learnable (shared)         & 2.6          \\ 
Learnable               & NLP            & 23.4      \\ 
NLP               & NLP            & 28.7       \\ 

\hline

\end{tabular}
\caption{\textbf{Comparison of NUS-WIDE mAP scores for ML-Decoder with different query types and head types}. "NLP" signifies using NLP projections to generate query/head parameters, and "Learnable" signifies using regular parameters (as described in Section~\ref{ml-decoder-design})} 

\label{Table:zsl_group_ablation}
\end{table}
\begin{table*}[hbt!]
\centering
\begin{tabular}{c|c|c|c|c|c|c} 
\Xhline{3\arrayrulewidth}
\begin{tabular}[c]{@{}c@{}}Classification\\Head\end{tabular}                       & \begin{tabular}[c]{@{}c@{}}Number of\\Classes\end{tabular} & \begin{tabular}[c]{@{}c@{}}Number of \\Queries\end{tabular} & \begin{tabular}[c]{@{}c@{}}Training\\Speed \\{[}img/sec]\end{tabular} & \begin{tabular}[c]{@{}c@{}}Inference\\Speed\\{[}img/sec]\end{tabular} & \begin{tabular}[c]{@{}c@{}}Maximal\\Training \\Batch Size\end{tabular}  & \begin{tabular}[c]{@{}c@{}}Flops \\{[}G]\end{tabular}  \\ 
\Xhline{3\arrayrulewidth}
\multirow{3}{*}{GAP}                                                          & 100                                                        & —                                                           & 706                                                                   & 2915                                                                  & 520                                                                    & 5.7    \\
                                                                              & 1000                                                       & —                                                           & 703                                                                   & 2910                                                                  & 512                                                                    & 5.7    \\
                                                                              & 5000                                                       & —                                                           & 698                                                                   & 2846                                                                  & 504                                                                    & 5.8    \\ 
\hline
\multirow{3}{*}{\begin{tabular}[c]{@{}c@{}}Transformer-\\Decode\end{tabular}} & 100                                                        & 100                                                         & 556                                                                   & 2496                                                                  & 424                                                                    & 6.6    \\
                                                                              & 1000                                                       & 1000                                                        & 44                                                                    & 916                                                                   & 112                                                                    & 14.2   \\
                                                                              & 5000                                                       & 5000                                                        & 2                                                                     & 17                                                                      & 4                                                                      & 61.1   \\ 
\hline
\multirow{3}{*}{ML-Decoder}                                                   & 100                                                        & 100                                                         & 575                                                                   & 2588                                                                  & 464                                                                    & 6.3    \\
                                                                              & 1000                                                       & 100                                                         & 568                                                                   & 2563                                                                  & 456                                                                    & 6.3    \\
                                                                              & 5000                                                       & 100                                                         & 562                                                                   & 2512                                                                  & 448                                                                    & 6.4    \\
\hline
\end{tabular}
\caption{\textbf{Comparison of throughput indices  for different classification heads.} All measurements were done on Nvidia V100 16GB machine, with mixed precision. We used TResNet-M as a backbone, with input resolution $224$. Training and inference speed were measured with 80\% of maximal batch size.}
\label{Table:througput_measurements}
\end{table*}


\section{MS-COCO Training Details}
\label{appendix:training_details_coco}

Unless stated explicitly otherwise, for MS-COCO we used the following training procedure:
We  trained  our  models for $40$ epochs using Adam optimizer  and 1-cycle policy, with maximal learning rate of $2\text{e-}4$.
For regularization, we used Cutout factor of $0.5$, True-weight-decay of $1\text{e-}4$ and auto-augment. We found that the common ImageNet statistics normalization does not improve results, and instead used a simpler normalization - scaling all the RGB channels to be between $0$ and $1$. Our input resolution was $448$. For ML-Decoder, our baseline was full-decoding ($K=N=80$).
Similar to \cite{ben2020asymmetric}, we used Open Images pretraining for our models' backbone. ML-Decoder weights were used with random initialization. The number of token embeddings was $D = 768$. We adjusted the backbone embedding output to $D$ via a $1\times1$ depth-wise convolution. As a loss function, we used ASL with $\gamma_{-}=4$, $\gamma_{+}=0$ and $m=0.05$. All results were averaged over three seeds for better consistency. TResNet-L model is V2\footnote{see: https://github.com/Alibaba-MIIL/TResNet}.

\section{Speed Comparison for Different Classification Heads}

In Table \ref{Table:througput_measurements} we provide full speed-accuracy measurements.
We can see that in practice, using ML-Decoder on TResNet-M architecture reduces inference speed by $15\%$, independent with the number of classes, while transformer-decoder classification head scales badly with the number of classes, reducing the inference speed (and other throughput metrices) by orders of magnitudes.

\section{MS-COCO Results for Different Input Resolutions}
\begin{table}[hbt!]
\centering
\begin{tabular}{c|c|c|c|c} 
\Xhline{3\arrayrulewidth}
Method     & Backbone  & \begin{tabular}[c]{@{}c@{}}Input \\Resolution\end{tabular} & \begin{tabular}[c]{@{}c@{}}Flops\\{[}G]\end{tabular} & \begin{tabular}[c]{@{}c@{}}mAP \\{[}\%]\end{tabular}  \\ 
\Xhline{3\arrayrulewidth}
ML-Decoder & TResNet-L & 224x224                                                    & 9.3                                                  & 85.5                                                  \\
ML-Decoder & TResNet-L & 448x448                                                    & 36.2                                                 & 90.0                                                  \\
ML-Decoder & TResNet-L & 640x640                                                    & 73.5                                                 & 91.1                                         \\
ML-Decoder & TResNet-XL & 640x640                                                    & 103.8                                                 & \textbf{91.4}                                         \\
\hline
\end{tabular}
\caption{\textbf{Comparison of MS-COCO mAP scores for different input resolutions}.}
\label{Table:coco_resolution_comparison}
\vspace{-0.2cm}
\end{table}

\section{NUS-WIDE ZSL Dataset and Training Details}
\label{appendix:training_details_nus_wide}
NUS-WIDE is a multi-label ZSL dataset, comprised of nearly $270K$ images with $81$ human-annotated categories, in addition to the $925$ labels obtained from Flicker user tags. The $925$ and $81$ labels are used as seen and unseen classes, respectively.

To be compatible with previous works~\cite{ben2021semantic}, as a loss function we used CE, our backbone was TResNet-M, and our input resolution is $224$. Unless stated otherwise, our baseline ML-Decoder for ZSL was full-decoding, with $K=N$, and shared projection matrix, as discussed in Section~\ref{scheme_zsl}.
Other training details for ZSL NUS-WIDE were similar to the one used for MS-COCO. 

\section{Pascal-VOC Training Details and Results}
Pascal Visual Object Classes
Challenge (VOC 2007) is another popular dataset for
multi-label recognition. It contains images from 20
object categories, with an average of 2.5 categories per image. Pascal-VOC is divided to a trainval set of 5,011 images and a test set of 4,952 images. As a backbone we used TResNet-L, with input resolution of $448$. For ML-Decoder, our baseline was full-decoding ($K=N=20$). Other training details are similar to the ones used for MS-COCO. Results appear in Table \ref{Table:pascal_voc}.
\begin{table}[hbt!]
\centering
\begin{tabular}{c|c} 
\Xhline{3\arrayrulewidth}
Method     & mAP [\%]   \\ 
\Xhline{3\arrayrulewidth}
RNN~\cite{wang2017multi}        & 91.9  \\
FeV+LV~\cite{yang2016exploit}     & 92.0  \\
ML-GCN~\cite{chen2019multi_MLGCN}     & 94.0  \\
SSGRL~\cite{chen2019learning}      & 95.0  \\
BMML~\cite{li2020bi}       & 95.0  \\
ASL~\cite{ben2020asymmetric}        & 95.8  \\
Q2L~\cite{liu2021query2label}        & 96.1  \\ 
\hline
ML-Decoder & 96.6 \\ 
\hline
\end{tabular}
\caption{\textbf{Comparison of ML-Decoder to known state-of-the-art models on Pascal-VOC dataset.} For ML-Decoder, we used TResNet-L backbone, input resolution $448$.}
\label{Table:pascal_voc}
\end{table}

\section{Open-Images Training Details and Results}
\label{appendix:training_details_open_images}
Open Images (v6) \cite{kuznetsova2018open} is a large-scale dataset, which consists of 9 million training images, $41,620$ validation images and  $125,436$ test images. It is partially annotated with human labels and machine-generated labels.
For dealing with the partial labeling methodology of Open Images dataset, we set all untagged labels as negative, with reduced weights. Due to the large the number of images, we trained our network for $25$ epochs on input resolution of $224$. We used TResNet-M as a backbone.
Since the level of positive-negative imbalancing is significantly higher than MS-COCO, we increased the level of loss asymmetry: For ASL, we trained with $\gamma_{-}=7, \gamma_{+}=0$. For ML-Decoder, our baseline was group-decoding with $K=100$.
Other training details are similar to the ones used for MS-COCO.
\begin{table}[hbt!]
\centering
\begin{tabular}{c|c} 
\Xhline{3\arrayrulewidth}
Method     & mAP [\%]   \\ 
\Xhline{3\arrayrulewidth}
CE~\cite{ben2020asymmetric}         & 84.8  \\
Focal Loss~\cite{ben2020asymmetric}     & 84.9  \\
ASL~\cite{ben2020asymmetric}            & 86.3  \\ 
\hline
ML-Decoder & 86.8 \\
\hline
\end{tabular}
\caption{\textbf{Comparison of ML-Decoder to known state-of-the-art results on Open Images dataset.}}
\label{Table:open_images}
\end{table}

\section{Single-label Classification with Different Logit Activations}

\begin{table}[hbt!]
\centering
\begin{tabular}{c|c|c} 
\Xhline{3\arrayrulewidth}
Logit Activation        & Classification Head & \multicolumn{1}{c}{Top1 Acc. [\%]}  \\ 
\Xhline{3\arrayrulewidth}
\multirow{2}{*}{Sigmoid} & GAP~           & 79.7                                 \\
                         & ML-Decoder     & 80.3                                 \\ 
\hline
\multirow{2}{*}{Softmax} & GAP~ ~         & 79.3                                 \\
                         & ML-Decoder     & 80.1                                 \\
\hline
\end{tabular}
\caption{\textbf{ImageNet classification scores for different classification heads and logits activations}. For ML-Decoder, we used group-decoding with 100 groups. Our training configuration is A2~\cite{wightman2021resnet}. Backbone - ResNet50.}
\label{Table:single_label_softmax_results}
\end{table}

\section{Comparison of ML-Decoder to State-of-the-art Models on Single-label Transfer Learning Datasets}
\label{appendix:additional_results}
In this section, we will compare our ML-Decoder based models to known state-of-the-art models from the literature, on two prominent single-label datasets - CIFAR-100\cite{cifar} and Stanford-Cars\cite{stanford_cars}. The comparison is based on \footnote{https://paperswithcode.com/sota/image-classification-on-cifar-100} and \footnote{https://paperswithcode.com/sota/fine-grained-image-classification-on-stanford}.

\begin{table}
\centering
\begin{tabular}{|c|c|c|} 
\hline
Dataset                                                                   & Model                     & \begin{tabular}[c]{@{}c@{}}Top-1\\Acc.\end{tabular}  \\ 
\hline\hline
\multirow{4}{*}{CIFAR-100}                                                & CvT-W24                   & 94.05                                                \\
                                                                          & ViT-H                     & 94.55                                                \\
                                                                          & EffNet-L2 (SAM)           & \textbf{96.08}                                       \\
                                                                          & Swin-L + ML-Decoder       & 95.1                                                 \\ 
\hline\hline
\multirow{4}{*}{\begin{tabular}[c]{@{}c@{}}Stanford-\\ Cars\end{tabular}} & EffNet-L2 (SAM)           & 95.95                                                \\
                                                                          & ALIGN                     & 96.13                                                \\
                                                                          & DAT                       & 96.2                                                 \\
                                                                          & TResNet-L + ML-Deocder & \textbf{96.41}                                       \\
\hline
\end{tabular}
\caption{\textbf{Comparison top of state-of-the-art models}.}
\label{Table:addional_datasets}
\end{table}
We can see from Table \ref{Table:addional_datasets} that our  ML-Decoder based solution is highly competitive, achieving 1st and 2nd place on Stanford-Cars and CIFAR-100 datasets respectivly.

\section{Query Augmentations Illustration}
\begin{figure}[hbt!]
    \centering
    \includegraphics[width=0.48\textwidth]{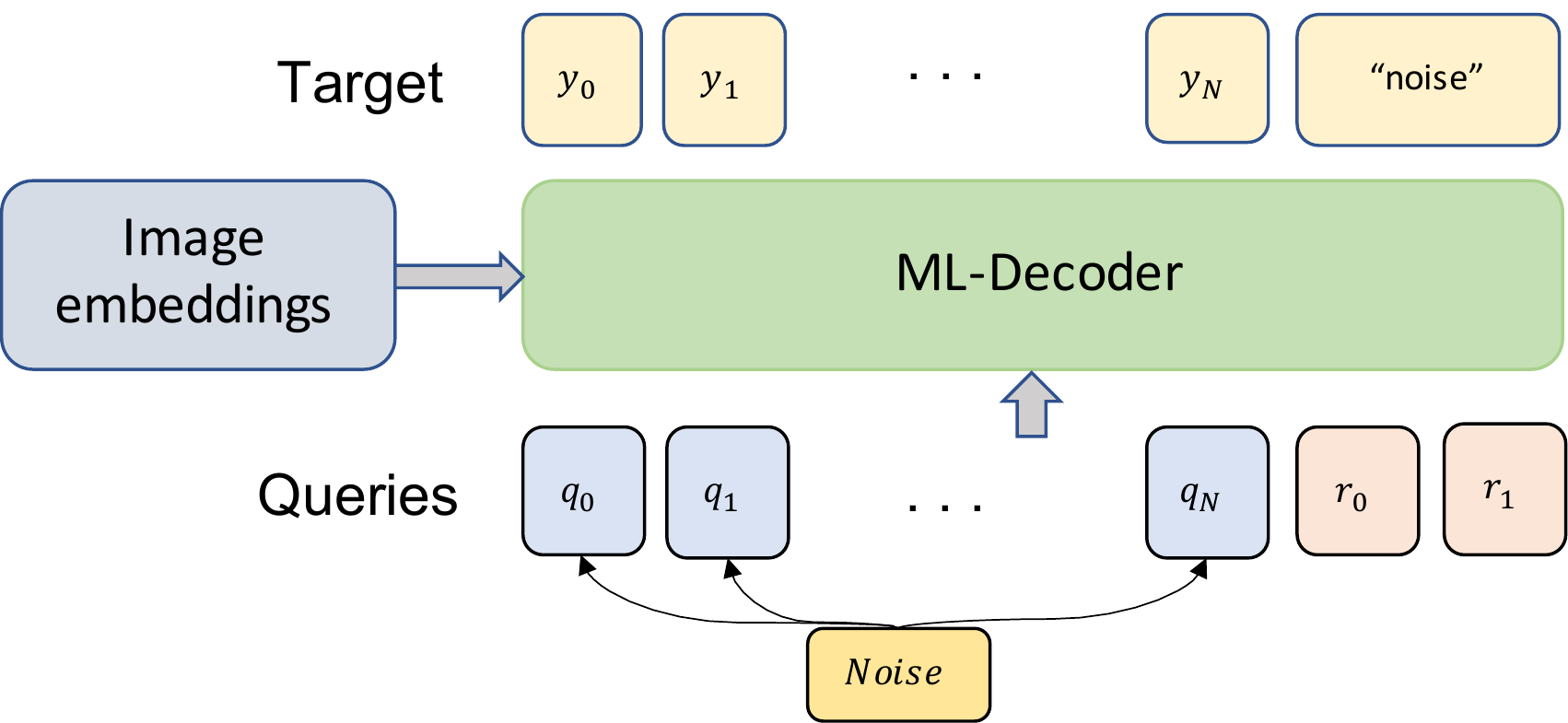}
    \caption{\textbf{Query augmentations} \emph{rand-query}: adding random queries which are assigned label "noise". \emph{additive noise}: adding random noise to the input queries.
    }
    \label{fig:query_aug}
    \vspace{-0.15cm}
\end{figure}

\clearpage
\section{Group Fully-connected Pseudo Code}
\label{appendix:group_pseudo_code}

\begin{lstlisting}
def GroupFullyConnected(G, group_weights, output, num_of_groups):
    '''
    - G is the group queries tensor. G.shape = [groups, embeddings]
    - group_weights are learnable (group) fully connected weights. 
      group_weights.shape = [groups, embeddings, classes//groups]
    - output is the interpolated queries tensor. output.shape = [groups, classes//groups]
    '''
    for i in range(num_of_groups):
        g_i = G[i, :] # [1,embeddings]
        w_i = group_weights[i, :, :] # [embeddings, classes//groups]
        output_i = matmul(g_i, w_i) # [1, classes//groups]
        output[i, :] = output_i
    logits = output.flatten(1)[:self.num_classes] # [1, classes]
    return logits
\end{lstlisting}

Notice that the loop implementation is very efficient in terms of memory consumption during training. Implementing the group-fully-connected in a single vectoric operation (without a loop) is possible, but reduces the possible batch size.
Also, the proposed implementation is fully suitable for compile-time acceleration (@torch.jit.script)

\section{Group Fully-connected ZSL Pseudo-Code}
\label{appendix:zsl_group_code}
\lstset{label={lst:code_zsl}}
\begin{lstlisting}
def GroupFullConnectedZSL(G, wordvecs, output, W, num_groups, num_classes):
    '''
    - G is the group queries tensor. G.shape = [num_groups, embeddings_dim]
    - wordvecs is the word-embedding tensor [num_classes, word_embedding_dim]
    - W is a learnable projection matrix [embeddings_dim, word_embedding_dim]
    - output is the interpolated queries tensor. output.shape = [num_groups, num_classes//num_groups]
    '''
    labels_per_group = num_classes // num_groups 
    group_weights = zeros(num_groups, embedding_dim, labels_per_group)
    for i in range(num_classes):
        group_weights[i // labels_per_group, :, i % labels_per_group] = W * wordvecs[i, :]
        
    logits = GroupFullyConnected(G, group_weights, output, num_groups)
    return logits
\end{lstlisting}

\end{appendices}

\end{document}